\documentclass[conference]{IEEEtran}

\IEEEoverridecommandlockouts
\usepackage{cite}
\usepackage{amsmath,amssymb,amsfonts}
\usepackage{algorithmic}
\usepackage{graphicx}
\usepackage{textcomp}
\usepackage{xcolor}
\usepackage{multirow}
\usepackage[squaren, Gray, cdot]{SIunits}
\usepackage{balance}

\def\BibTeX{{\rm B\kern-.05em{\sc i\kern-.025em b}\kern-.08em
    T\kern-.1667em\lower.7ex\hbox{E}\kern-.125emX}}

\usepackage[lofdepth,lotdepth]{subfig}
\usepackage{tikz}
\usepackage{pgfplots}
\usepackage{filecontents}
\usepackage[skins]{tcolorbox}

\usetikzlibrary{positioning, fit, calc, arrows.meta}
\usepgfplotslibrary{statistics}

\newcommand{\falez}[1]{#1}
\newcommand{\tirilly}[1]{#1}
\newcommand{\marius}[1]{#1}

\begin{document}

\title{Multi-layered Spiking Neural Network with Target Timestamp Threshold Adaptation and STDP
\thanks{This work has been partly funded by IRCICA (Univ. Lille, CNRS,
USR 3380 – IRCICA, F-59000 Lille, France) as the Bioinspired Project.}
}

\DeclareRobustCommand*{\IEEEauthorrefmark}[1]{%
  \raisebox{0pt}[0pt][0pt]{\textsuperscript{\footnotesize #1}}%
}

\author{
\IEEEauthorblockN{
Pierre Falez\IEEEauthorrefmark{1},
Pierre Tirilly\IEEEauthorrefmark{1}\,\IEEEauthorrefmark{2},
Ioan Marius Bilasco\IEEEauthorrefmark{1},
Philippe Devienne\IEEEauthorrefmark{1},
and Pierre Boulet\IEEEauthorrefmark{1}
}
\IEEEauthorblockA{\IEEEauthorrefmark{1}
\textit{Univ. Lille, CNRS, Centrale Lille,}
\textit{UMR 9189 -- CRIStAL -- Centre de Recherche en Informatique, Signal et Automatique de Lille}\\
F-59000, Lille, France\\
}
\IEEEauthorblockA{\IEEEauthorrefmark{2}
\textit{IMT Lille Douai, F-59000, Lille, France}
}
\IEEEauthorblockA{
Email: firstname.lastname@univ-lille.fr
}
}

\maketitle

\begin{abstract}
Spiking neural networks (SNNs) are good candidates to produce ultra-energy-efficient hardware. However, the performance of these models is currently behind traditional methods. Introducing multi-layered SNNs is a promising way to reduce this gap. We propose in this paper a new threshold adaptation system which uses a timestamp objective at which neurons should fire. We show that our method leads  to state-of-the-art classification rates on the MNIST dataset (98.60\%) and the Faces/Motorbikes dataset (99.46\%) with an unsupervised SNN followed by a linear SVM. We also investigate the sparsity level of the network by testing different inhibition policies and STDP rules.
\end{abstract}


\section{Introduction}

Computer vision has rapidly evolved in recent years, in particular thanks to deep learning methods~\cite{lecun2015deep}. They show that using deep hierarchical representations improves the expressiveness of models~\cite{montufar2014number}, and yields state-of-the-art performance on many tasks~\cite{he2015delving}~\cite{silver2017mastering}. However, the question of the energy consumption of such models remains less frequently addressed, even though it has been raised by some authors~\cite{bekkerman2011scaling}~\cite{cao15a}~\cite{hubara18a}.

Although efforts are being made to produce more energy-efficient architectures for traditional methods~\cite{shan_tang_deep-learning-processor-list_nodate}, producing ultra-low-power architectures seems to require using different classes of models. Spiking neural networks (SNNs) are good candidates to create energy-efficient hardware~\cite{merolla2014million}~\cite{sourikopoulos20174}. To achieve this goal, SNNs use mechanisms closer to biology, notably the fact that computations and memory are exclusively local~\cite{paugam2012computing}. Instead of using numerical representations like traditional methods, SNNs use spikes to transmit information, which radically changes their learning mechanisms. However, their classification performances currently remain behind traditional deep learning methods~\cite{falez2018}. This gap is in part due to the constraint of local computation in SNNs, which prevents the use of traditional training methods like back-propagation. New learning mechanisms are necessary to bypass this limit and allow them to compete with state-of-the-art methods.

The performance of SNN is highly sensitive to network meta-parameters. As a consequence, an exhaustive search for all parameters, which are numerous, is generally required when using SNNs. This makes SNNs difficult to use. Reducing the impact of parameter values, by using auto-adaptive parameters, or at least, reducing the number of parameters, seems to be a key point in order to be able to make SNNs viable.

More specifically, neuron thresholds are one of the key parameters in a SNN. They determine the amount of spikes required by the neurons to trigger a spike; it directly impacts the patterns that neurons can recognize, and so, the performance of the network. The optimal threshold value can vary widely over the different neurons in a network because the inputs or internal patterns are made up of different numbers of spikes.

In this paper, we propose a new threshold learning rule, which is based on a target timestamp $t_{\sf target}$ at which a neuron must fire. This target timestamp directly controls the patterns that neurons can learn. By providing an adaptive threshold, this mechanism reduces the impact of its initial value. Moreover, thanks to the usage of a unique parameter, the search space to optimize is relatively small. Additionally, we provide a protocol to train multi-layered networks. We evaluate this mechanism with multi-layered SNNs on the Faces/Motorbikes~\cite{kheradpisheh2018stdp} and MNIST~\cite{lecun1998gradient} datasets. We study the impact of our threshold adaptation system, but also of the inhibition policy and of the STDP rule. Finally, we show that we can combine multiple networks trained with different $t_{\sf target}$ to improve the classification rate thanks to the different patterns learned by the network. This method reaches state-of-the-art results on both the Faces/Motorbikes and MNIST datasets.

\section{Related work}

Early work in image recognition with fully unsupervised SNNs used single-layered networks~\cite{querlioz2011simulation}~\cite{diehl2015unsupervised}. However, such methods yield low classification rates on the MNIST dataset  (93.5\%~\cite{querlioz2011simulation}, 95\%~\cite{diehl2015unsupervised}) compared to traditional methods~\cite{lecun1998gradient}. One of the first multi-layered STDP networks is~\cite{tavanaei2016bio}. In this work, a dedicated network, SAILNet, learns convolution filters from patches extracted from input samples.
A pooling layer and a fully connected layer using probabilistic LIF neurons are stacked. A support vector machine (SVM) classifies the output of the last layer. This model reaches 98.36\% on MNIST with 32 convolution filters and 128 output neurons. However, the usage of an external network to train convolutions remains an issue. Moreover, probabilistic LIF neurons are used in the feature discovery layer, which requires some global computation (softmax) to operate.

In \cite{kheradpisheh2018stdp}, two convolution layers trained by STDP are used. The network reaches 98.4\% on the MNIST dataset with 30 filters in the first convolution layer and 100 in the second one. However, this model uses some global computations: the potentials of neurons are compared to each other to designate the winner at every step and the filters are learned across the convolution columns. This model requires to tune its parameters carefully, especially the neuron thresholds. Moreover, the values of neuron thresholds must be manually changed between the training and testing stages. Finally, the output neurons use infinite thresholds, which would not be realistic on hardware.

Other authors focus on converting traditional deep neural networks, trained with back-propagation, into multi-layered SNNs~\cite{diehl2015fast}~\cite{cao2015spiking}. However, this method limits the interest of SNNs, since only the inference stage can be energy-efficient. Moreover, such networks are not able to adapt themselves continuously, since their parameters are fixed after the conversion. Other work adapts the back-propagation method to SNNs~\cite{bohte2002error}~\cite{o2017temporally}~\cite{bengio2015towards}. However, these models cannot be as energy-efficient since they need global computations to perform back-propagation.

\section{Background}

\begin{figure}
    \centering
    \resizebox{0.9\columnwidth}{!}{\begin{tikzpicture}[
	node distance=1mm,
	layer/.style={black, draw=black, dashed},
	neuron/.style={black, draw=black, circle, minimum size=0.75cm,inner sep=0.5pt},
	preconnection/.style={shorten >=1pt, shorten <=1pt, >=stealth},
	postconnection/.style={-{Latex}, shorten >=1pt, shorten <=1pt, >=stealth},
	synapse/.style={black,draw=black,minimum width=1cm,minimum height=0.5cm},
	spike/.pic={
    \draw () -- ++(#1,0) -- ++(0,0.2mm);
},
	spike2/.pic={
     \draw[black, thick] (0,0) -- ++(0,2mm);
}
]

	\node (layer_input_ref) {};

	\node[neuron] (neuron_1_3)  at (layer_input_ref) {$I_3$};
	\node[neuron] (neuron_1_2) [above = of neuron_1_3] {$I_2$};
	\node[neuron] (neuron_1_1) [ above = of neuron_1_2] {$I_1$};
	\node[neuron] (neuron_1_4) [ below = of neuron_1_3] {$I_4$};
	\node[neuron] (neuron_1_5) [ below = of neuron_1_4] {$I_5$};
	\node (layer_1) [fit={(neuron_1_1)  (neuron_1_2) (neuron_1_3) (neuron_1_4) (neuron_1_5)}] {};

	\node (neuron) [neuron, right = 40mm of layer_1]  {$N$};

	\coordinate (input_end) at ($(neuron)+(-15mm,0)$);

	\node[synapse]  (synapse_1_1) at (neuron_1_1 -| input_end) {$S$};
	\node[synapse]  (synapse_1_2) at (neuron_1_2 -| input_end) {$S$};
	\node[synapse]  (synapse_1_3) at (neuron_1_3 -| input_end) {$S$};
	\node[synapse]  (synapse_1_4) at (neuron_1_4 -| input_end) {$S$};
	\node[synapse]  (synapse_1_5) at (neuron_1_5 -| input_end) {$S$};

	\draw [preconnection] (neuron_1_1)  -- (synapse_1_1) ;
	\draw [preconnection] (neuron_1_2)  -- (synapse_1_2);
	\draw [preconnection] (neuron_1_3)  -- (synapse_1_3);
	\draw [preconnection] (neuron_1_4)  -- (synapse_1_4);
	\draw [preconnection] (neuron_1_5)  -- (synapse_1_5);

	\draw [postconnection] (synapse_1_1.east) -- (neuron);
	\draw [postconnection] (synapse_1_2.east) -- (neuron);
	\draw [postconnection] (synapse_1_3.east) -- (neuron);
	\draw [postconnection] (synapse_1_4.east) -- (neuron);
	\draw [postconnection] (synapse_1_5.east) -- (neuron);

	\node (output_layer)  [right = 50mm of neuron] {};

	\node[neuron] (neuron_2_2)  at (output_layer) {$O_2$};
	\node[neuron] (neuron_2_1) [above = of neuron_2_2] {$O_1$};
	\node[neuron] (neuron_2_3) [ below = of neuron_2_2] {$O_3$};

	\coordinate (output_end) at ($(output_layer)+(-25mm,0)$);
	\coordinate (synapse_out) at ($(output_layer)+(-15mm,0)$);

	\node[synapse]  (synapse_2_1) at (neuron_2_1 -| synapse_out) {$S$};
	\node[synapse]  (synapse_2_2) at (neuron_2_2 -| synapse_out) {$S$};
	\node[synapse]  (synapse_2_3) at (neuron_2_3 -| synapse_out) {$S$};
	
	\draw(neuron) --  (output_end);

	\draw[preconnection] (output_end) --  (synapse_2_1.west);
	\draw[preconnection] (output_end) --  (synapse_2_2.west);
	\draw[preconnection] (output_end) --  (synapse_2_3.west);

	\draw[postconnection] (synapse_2_1) --  (neuron_2_1);
	\draw[postconnection] (synapse_2_2) --  (neuron_2_2);
	\draw[postconnection] (synapse_2_3) --  (neuron_2_3);

	\pic at  ($(neuron_1_1)+(10mm,0)$) {spike2};
	\pic at  ($(neuron_1_1)+(13mm,0)$) {spike2};
	\pic at  ($(neuron_1_1)+(26mm,0)$) {spike2};

	\pic at   ($(neuron_1_2)+(20mm,0)$) {spike2};

	\pic at ($(neuron_1_4)+(5mm,0)$){spike2};
	\pic at ($(neuron_1_4)+(12mm,0)$) {spike2};
	\pic at ($(neuron_1_4)+(25mm,0)$) {spike2};

	\pic at ($(neuron_1_5)+(20mm,0)$) {spike2};
	\pic at ($(neuron_1_5)+(23mm,0)$){spike2};

	\pic at  ($(neuron)+(15mm,0)$) {spike2};
	\pic at   ($(neuron)+(23mm,0)$) {spike2};

	\coordinate(axis) at(0,-25mm);
	\draw[-{Latex}] (layer_1.east |- axis) -- node[below]{t} (synapse_1_1.west|- axis);
	\draw[-{Latex}] (neuron.east |- axis) -- node[below]{t} (output_end|- axis);

\end{tikzpicture}}
    \caption{A spiking neuron receives spikes from a set of input neurons and generates spikes towards a set of output neurons. On each connection, a synapse modulates the spike voltage.}
    \label{fig:snn}
\end{figure}
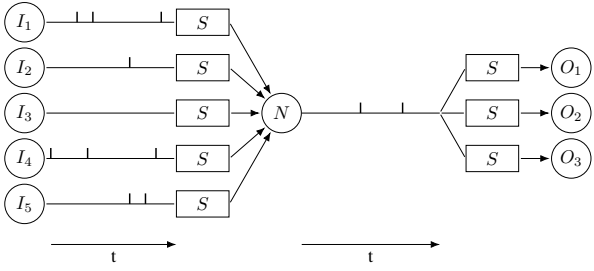

In contrast to traditional artificial neural networks (ANNs), which use numerical values to represent information, SNNs use electrical impulses, called spikes. In this paper, for simplicity reasons, spikes are represented by a Dirac impulsion, defined by a timestamp $t$ and a voltage $V$ (Figure~\ref{fig:snn}).

\subsection{Pre-processing}

Before input samples are converted into spikes to be fed to the network, some pre-processing steps are applied. We use a difference-of-Gaussians (DoG) filter to simulate on-off cells~\cite{delorme2001networks}. Without this pre-processing, SNNs fail to learn useful patterns, leading to low classification performances~\cite{falez2018}.

DoG filters are applied with the same process as the one described in~\cite{kheradpisheh2018stdp}:
\begin{align}
	&\mathrm{DoG}(x, y) = I(x,y) \ast (G_{\mathrm{DoG}_{\sf size}, \mathrm{DoG}_{\sf center}} - G_{\mathrm{DoG}_{\sf size}, \mathrm{DoG}_{\sf surround}}) \nonumber
\end{align}
where $I$ is the input image, \(\ast\) is the convolution operator and \(G_{K, \sigma}\) is a normalized Gaussian kernel of size \(K\) and scale \(\sigma\) defined as:
\begin{align}
	&G_{K, \sigma}(u,v) =\frac{g_\sigma(u, v)}{\sum\limits_{i = -\mu}^\mu \sum\limits_{j = -\mu}^\mu g_\sigma(i, j)}, u, v \in [-\mu, \mu], \mu = \frac{K}{2}, \nonumber
\end{align}
with $g_\sigma$ the centered 2D Gaussian function of variance \(\sigma\). The parameters of the filter are its size $\mathrm{DoG}_{\sf size}$ and the variances of the Gaussian kernels $\mathrm{DoG}_{\sf center}$ and $\mathrm{DoG}_{\sf surround}$.

After applying DoG filtering over an input image, the resulting values are separated into two channels:
\begin{equation}
    \begin{aligned}
     x_{\sf on} & = \max(0, \mathrm{DoG}(x,y))\\
     x_{\sf off} & = \max(0, -\mathrm{DoG}(x,y))
   \end{aligned}
\end{equation}

\subsection{Neural Coding}

Since SNNs use spikes to transfer information within the network, it is necessary to define a function to encode the numerical values of input samples into spikes trains and a function to decode spike trains at the output of the network. The encoding function is referred to as the neural coding. Mathematically, a neural coding can be described as follows:
\begin{equation}
   \begin{aligned}
     f: [0, 1] & \rightarrow \mathbb{R}_+^{N_x}\\
     x & \mapsto \left(t_0, t_1, \cdots, t_{N_x}\right)
   \end{aligned}
\end{equation}
with $x$ the input pixel value and $\left(t_0, t_1, \cdots, t_{N_x}\right)$ the \mbox{timestamps} of the generated spikes.

Neural coding is subject to debate in the SNN community~\cite{brette2015philosophy}. Two main coding \marius{techniques} exist: frequency coding and temporal coding. While frequency coding uses spike frequencies to encode values, temporal coding uses the timestamps of spikes. One of the most used methods is latency coding~\cite{thorpe2001spike}, in which early spikes encode the largest values, while late spikes encode the lowest values:
\begin{equation}
  t = T_{\sf start}+(1-x)*(T_{\sf end}-T_{\sf start})
\end{equation}
with $[T_{\sf start}, T_{\sf end}]$ the time range of the sample, $x \in [0, 1]$ the input value, and $t$ the timestamp of the generated spike.

This paper uses latency coding as neural coding as it has the main advantage of using few spikes (at most one spike per connection) to represent values, which makes the model easier to control. However, in latency coding, the timestamps at which neurons discharge are critical since they have a direct impact on the represented values.

\subsection{Neuron Model}

In this paper, we use integrate-and-fire (IF) neurons, which are one of the simplest spiking neuron models. This model integrates input spikes to its membrane potential $V$. If $V$ exceeds a defined threshold  $V_{\sf th}$, then an output spike is triggered and $V$ is reset to 0. The model is defined by the following formula:

\begin{equation}
	\frac{\partial V}{\partial t} = \sum\limits_{i\in S} V_i\delta(t-t_i), V \leftarrow 0~{\sf when}~V \geq V_{\sf th}
\end{equation}

with $S$ the set of incoming spikes, $V_i$ the voltage of the $i$\textsuperscript{th} spike, $t_i$ the timestamp of the $i$\textsuperscript{th} spike and $\delta$ the Dirac function. In addition, all potentials are reset to zero between each sample. If a neuron fires a spike during the presentation of a sample, it enters its refractory mode until the end of the sample. This constraint forces neurons to fire at most once per sample, in order to comply with latency coding.

\subsection{Synapse Model}
\label{ssec:synapses}

Synapses modulate the spike voltage $V$ that passes through connections according to their synaptic weights $W$: $V_{O} = WV_{I}$, with $V_{I}$ the voltage of the spike at the input of the synapse and $V_{O}$ its voltage at the output. This weight can be constant or can be trained following a learning rule. In our synapse model, $W$ is clipped in the range [$W_{\sf min}$, $W_{\sf max}$]. One of the most used learning rules is spike-timing-dependent plasticity (STDP)~\cite{bi1998synaptic}, which updates the weights according to the difference between the firing timestamps of the input neuron and the output neuron. One of the simplest forms of the STDP rule is additive STDP~\cite{bichler2011unsupervised}. Its principle is to increase connection weights where input neurons fire spikes before output neurons (long-term potentiation) and to decrease the others (long-term depression). Mathematically, additive STDP can be written as:
\begin{equation}
    \Delta_W = \left \{
   \begin{array}{r l}
     \eta_W & \text{if } t_{\sf pre} \leq t_{\sf post} \\
     -\eta_W & \text{o.w.}
   \end{array}
   \right .
\end{equation}
with $\Delta_W$ the weight variation,  $\eta_W$ the learning rate, $t_{\sf pre}$ the firing timestamp of the input neuron ($+\infty$ if no spike occurs) and $t_{\sf post}$ the firing timestamp of the output neuron.

Other forms of STDP exist in the literature. Multiplicative STDP~\cite{querlioz2011simulation} allows reducing the effect of weight saturation by using updates that depend on the current value of $W$. This STDP rule is defined by the following formula:
\begin{equation}
    \Delta_W = \left \{
   \begin{array}{r l}
     \eta_W e^{-\beta\frac{W-W_{\sf min}}{W_{\sf max}-W_{\sf min}}} & \text{if }  t_{\sf pre} \leq t_{\sf post} \\
     -\eta_W e^{-\beta\frac{W_{\sf max}-W}{W_{\sf max}-W_{\sf min}}} & \text{o.w.}
   \end{array}
   \right .
\end{equation}
with $\beta$ the parameter which controls the saturation effect (increasing $\beta$ reduces the saturation).

Finally, biological STDP~\cite{bi1998synaptic} adds non-linearity by including a leak according to the delay between $t_{\sf pre}$ and $t_{post}$:
\begin{equation}
    \Delta_W = \left \{
   \begin{array}{r l}
     \eta_W e^{-\frac{t_{\sf pre}-t_{\sf post}}{\tau}} & \text{if } t_{\sf pre} \leq t_{\sf post} \\
     -\eta_W e^{-\frac{t_{\sf post}-t_{\sf pre}}{\tau}} & \text{o.w.}
   \end{array}
   \right .
\end{equation}
with $\tau$ the time constant that controls the leak.

\subsection{Network Architecture}

The network is composed of stacked feed-forward layers. For a layer $L(n)$, there are $L_d(n)$ feature maps, each of them containing $L_w(n) \times L_h(n)$ neurons. Three types of layers are used in this paper: convolution, pooling, and fully-connected layers. The shape of a layer depends on its filter size $F_w(n) \times F_h(n)$, its padding $P(n)$, and its stride $S(n)$. Each neuron of layer $L(n)$ is connected to $F_w(n-1) \times F_h(n-1) \times L_d(n-1)$ neurons of the previous layer, which form the receptive field of the neuron. In the pooling layers, all the parameters are constant: neuron thresholds and synaptic weights are fixed to 1. When a spike is triggered in its receptive field, a pooling neuron directly fires a spike. This mimics a max-pooling operation. A column $C_{x,y}(n)$ designates the $L_d(n)$ neurons present at position $(x, y)$ in the $L_d(n)$ features maps of $L(n)$.

\section{Contribution}

The introduction of new mechanisms to help neurons fire spikes at optimal timestamps is of paramount importance. We introduce in this paper a novel threshold adaption mechanism that trains neurons to discharge at a defined timestamp.

\subsection{Time target threshold adaptation}

\begin{figure}
    \centering
    \subfloat[0.4]{
        \includegraphics[width=0.2\columnwidth]{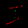}
    }
    \subfloat[0.6]{
        \includegraphics[width=0.2\columnwidth]{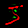}
    }
    \subfloat[0.8]{
        \includegraphics[width=0.2\columnwidth]{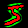}
    }
    \subfloat[1.0]{
        \includegraphics[width=0.2\columnwidth]{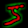}
    }
    \caption{Examples of learned patterns according to the time threshold. Only the spikes integrated before the threshold is reached are used in the pattern recognition process.}
    \label{fig:input_ex}
\end{figure}

Thresholds have a major role in the behavior of spiking neurons~\cite{carlson2013biologically}. First, threshold values directly impact the patterns recognized by neurons. Large threshold values will allow \marius{recognizing} patterns composed of large numbers of spikes (Figure~\ref{fig:input_ex}).
Neurons with smaller threshold values will use only the first input spikes in the pattern recognition process. Since latency coding is used, the first spikes encode the largest values of the input sample, which means that the neuron focuses on the most salient parts of the input, learning very local patterns, like edges. On the contrary, large threshold values allow the neuron to integrate more spikes before firing, including late spikes, which encode smaller values. Neurons with large thresholds can recognize larger patterns, like surfaces. However, the optimal threshold value is unknown and is highly dependent on the data. Finally, threshold adaptation allows \marius{maintaining} the homeostasis of the system: it ensures that no neuron takes advantage over the others. A common method to adapt thresholds in SNNs is to use leaky adaptive thresholds~\cite{diehl2015unsupervised}: when a neuron fires a spike, its threshold is increased to prevent it from firing too often. An exponential leak is applied to help neurons with weak activities. However, this mechanism uses two parameters, which makes the search of suited values difficult~\cite{falez2018mastering}. Moreover, those parameters do not \marius{enhance the convergence} towards the different types of patterns shown in Figure~\ref{fig:input_ex}. This paper introduces a new method to adjust neuron thresholds. The idea is to define an objective timestamp $t_{\sf target}$, and to train neurons to fire at this timestamp. To do so, we define a threshold adaptation rule, as follows:
\begin{equation}
     V_{\sf th} = \max(\mathrm{Th}_{\sf min}, V_{\sf th}-\eta_{\sf th}(t-t_{\sf target}))
     \label{eq:th_target}
\end{equation}
with $V_{\sf th}$ the neuron threshold, $t$ the timestamp at which the neuron fires, $\eta_{th}$ the threshold learning rate and $\mathrm{Th}_{\sf min}$  the minimal threshold allowed. This rule corrects the timing error between the actual firing timestamp $t$ and the objective timestamp $t_{\sf target}$ at each neuron discharge. The optimal value for $t_{\sf target}$ depends on the dataset; it requires an exhaustive search in the range $[T_{\sf start}, T_{\sf end}]$.

\tirilly{This rule assumes that the input spikes that trigger an output spike are not simultaneous, which is the case in practice with image data. With data that does not verify this assumption, synaptic delays would have to be adapted.}

\subsection{Competition System}
\label{ssec:competition}

Using local and unsupervised learning requires competition mechanisms in order to ensure that neurons learn distinct patterns~\cite{querlioz2011simulation}. Winner-take-all (WTA) inhibition is a straightforward method to do so: only the winning neuron (i.e. the first neuron to spike, since latency coding is used) will apply the learning rule during a pattern and, so, will be able to recognize it. However, the risk of the WTA strategy is that one neuron can take the advantage over the others, and win on every sample. To guarantee the homeostasis of the system, a second update is applied to $V_{\sf th}$: the threshold of the winning neuron is increased, while the thresholds of inhibited neurons are decreased, following the formula:

\begin{equation}
\begin{aligned}
    \Delta V_{{\sf th}_i} &= \left\{
   \begin{array}{r l}
    \eta_{\sf th} & \text{if }t_i=\min\{t_0,\cdots,t_N\}\\
   -\frac{\eta_{\sf th}}{N} & {\sf o.w.}
   \end{array}
   \right .\\
    V_{{\sf th}_i} &= \max(\mathrm{Th}_{\sf min},  V_{{\sf th}_i}+\Delta V_{{\sf th}_i})
\end{aligned}
\label{eq:th_wta}
\end{equation}

with $N$ the number of neurons in competition, and $t_i$ the firing timestamp of neuron $i$.

WTA inhibition is used during training: only one neuron is allowed to fire among the $N$ neurons on each sample. This mechanism is required to guarantee that neurons will learn different patterns, since only one neuron will apply STDP per sample. However, WTA inhibition drastically reduces the spiking activity, which can lead to poor classification performance~\cite{falez2018}. For this reason, we remove the inhibition mechanism during the inference stage. An intermediate inhibition policy, named soft inhibition, is also investigated in this paper. This policy uses inhibition spikes, which reduce the membrane voltage $V$ of the other neurons by a $V_{\mathrm{inh}}$ constant, but does not prevent them from firing spikes.

\subsection{Network Output}

The last step is to interpret the output of the network. Since latency coding is used, the earliest output spikes will encode the highest values. Output values are computed according to the expected $t_{\sf target}$ set in the output layer, as follows:

\begin{equation}
    x = \min\left(1, \max\left(0, 1-\frac{t-t_{\sf target}}{T_{\sf end}-t_{\sf target}}\right)\right)
\end{equation}
with $t$ the spike timestamp (set to $+\infty$ if no spike occurs).

\subsection{Training}
\label{ssec:training}

Traditionally, convolution layers require to perform non-local operations and to use non-local memory since they use shared weights: columns need to communicate with each other to share the same filters. We use a specific training protocol in order to reduce the cost of the global communication needed by the convolutions. One layer is trained at a time, from the layer closest to the input to the one at the output of the network. During the training of a convolution layer, only one column is activated to avoid the usage of inter-column communications. Once the layer is trained, its parameters (weights and thresholds) are fixed and are copied onto the other columns of the layer. This operation is necessary since pooling layers require the same filters in adjacent columns. In order to keep the position invariance brought by shared weights, random patches of size $F_w(n) \times F_h(n)$ are extracted from inputs of the layer. Unlike in~\cite{kheradpisheh2018stdp}, neurons do not react only to the most salient part of each image.

\section{Results}

\subsection{Experimental Protocol}

For each trained layer, the training set is processed $N_{\sf epoch}$ times. A simulated annealing procedure is applied after every epoch: the learning rates ($\eta_W$ and $\eta_{Th}$) are decreased by a factor $\lambda$. It helps converge to a stable state during training. Once the training is finished, the training set and the test set are processed by the network, which converts all the samples into their output representation. If the output layer has multiple columns (i.e. $L_w(n) > 1$ or $L_h(n) > 1$), sum pooling is applied over the positions of the feature maps to produce a feature vector $y  = (y_1, ..., y_x)$:
\begin{equation}
    y_i = \sum\limits_{x=0}^w\sum\limits_{y=0}^h v_{xyi}
\end{equation}
with $v_{xyi}$ the value of output of the network at position $(x, y)$ in feature map $i$.

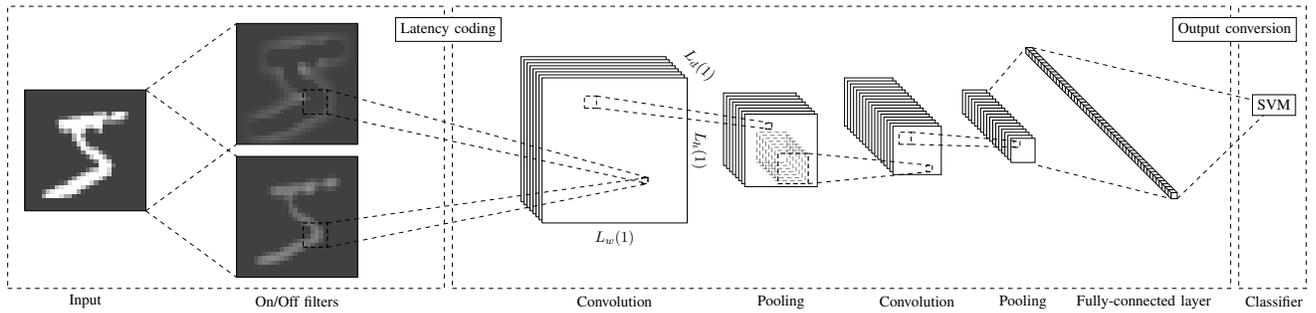
\begin{figure*}
    \centering
    \resizebox{\textwidth}{!}{\newcommand{\feature}[2]
{ 
	\draw[fill=white] ($(#1)+(-#2,-#2)$) -- ($(#1)+(#2,-#2)$) -- ($(#1)+(#2,#2)$) -- ($(#1)+(-#2,#2)$) --  ($(#1)+(-#2,-#2)$) ;
}

\newcommand{\rect}[3]
{ 
	\draw[fill=white] ($(#1)+(-#2,-#3)$) -- ($(#1)+(#2,-#3)$) -- ($(#1)+(#2,#3)$) -- ($(#1)+(-#2,#3)$) --  ($(#1)+(-#2,-#3)$) ;
}

\newcommand{\layer}[4]
{ 
	\foreach \x in {#3,...,0}{
		\feature{$(#1)+(-\x*#4,\x*#4)$}{#2}
	}
}

\newcommand{\denseLayer}[3]
{ 
	\draw[fill=white] ($(#1)+(-\s,-\s)$) -- ($(#1)+(\s,-\s)$) -- ($(#1)+(\s,\s)$) -- ($(#1)+(-\s,\s)$) --  ($(#1)+(-\s,-\s)$) ;

	\foreach \x in {#2,...,0}{
			\draw[fill=white] ($(#1)+(-\x*#3-\s,\x*#3-\s)$) -- ($(#1)+(-\x*#3-\s,\x*#3+\s)$) -- ($(#1)+(-\x*#3+\s,\x*#3+\s)$) ;
	}

	\draw[fill=white] ($(#1)+(-#2*#3-\s,#2*#3-\s)$) -- ($(#1)+(-#2*#3-\s,#2*#3+\s)$) -- ($(#1)+(-#2*#3+\s,#2*#3+\s)$) ;

	\draw[fill=white] ($(#1)+(-\s,-\s)$) -- ($(#1)+(-#2*#3-\s,#2*#3-\s)$);
	\draw[fill=white] ($(#1)+(-\s,\s)$) -- ($(#1)+(-#2*#3-\s,#2*#3+\s)$);
	\draw[fill=white] ($(#1)+(\s,\s)$) -- ($(#1)+(-#2*#3+\s,#2*#3+\s)$);

}

\newcommand{\convfeature}[3]
{ 
	\draw[densely dashed,opacity=#3] ($(#1)+(-#2,-#2)$) -- ($(#1)+(#2,-#2)$) -- ($(#1)+(#2,#2)$) -- ($(#1)+(-#2,#2)$) --  ($(#1)+(-#2,-#2)$) ;
}

\newcommand{\conv}[7]
{
    \conva{#1}{#2}{#3}{#4}{#5}{#6}{#7}
    \convb{#1}{#2}{#3}{#4}{#5}{#6}{#7}
}
\newcommand{\conva}[7]
{ 
	\foreach \x in {#3,...,0}{
		\convfeature{$(#1)+(-\x*#7,\x*#7)$}{#2}{0.5}
	}
}

\newcommand{\convb}[7]
{ 
	\convfeature{$(#1)+(0,0)$}{#2}{1.0}
	\convfeature{$(#4)+(0,0)$}{#5}{1.0}
	\draw[dashed] ($(#1)+(#2,-#2)$) -- ($(#4)+(#5,-#5)$) ;
	\draw[dashed] ($(#1)+(#2,#2)$) -- ($(#4)+(#5,#5)$) ;
}

\newcommand{\poolfeature}[2]
{ 
	\draw[fill=white, densely dashed] ($(#1)+(-#2,-#2)$) -- ($(#1)+(#2,-#2)$) -- ($(#1)+(#2,#2)$) -- ($(#1)+(-#2,#2)$) --  ($(#1)+(-#2,-#2)$) ;
}

\newcommand{\pool}[7]
{ 
	\convfeature{$(#1)+(0,0)$}{#2}{1.0}
	\convfeature{$(#4)+(0,0)$}{#5}{1.0}
	\draw[dashed] ($(#1)+(#2,-#2)$)  -- ($(#4)+(#5,-#5)$) ;
	\draw[dashed] ($(#1)+(#2,#2)$) -- ($(#4)+(#5,#5)$) ;
}

\newcommand{\dense}[6]
{ 
	\draw[dashed] ($(#1)+(-#2*#5+#6*#5+2*#5-2*#5,#2*#5+#6*#5+#5)$) -- ($(#3)+(-#4*#5-#5,#4*#5-#5+2*#5)$) ;
	\draw[dashed] ($(#1)+(#6*#5+#5, -#6*#5-#5)$)  -- ($(#3)+(-#5,-#5)$) ;
}

\begin{tikzpicture}
    \pgfgettransformentries{\a}{\b}{\c}{\d}{\xtrans}{\ytrans}
	
	\newcommand{\s}{0.07}
	\newcommand{\ca}{0}
	\newcommand{\cb}{7}
	\newcommand{\cc}{15}
	\newcommand{\cd}{48}

	\newcommand{\converterw}{10}
	\newcommand{\converterh}{5}

	\coordinate (input) at (0, 0);
	\layer{input}{20*\s}{\ca}{\s}

    \coordinate (oncell) at (70*\s, -22*\s);
	\layer{oncell}{20*\s}{0}{\s}

 	\coordinate (offcell) at (70*\s, 22*\s);
	\layer{offcell}{20*\s}{0}{\s}

	\coordinate (inconverter) at (120*\s, 40*\s);

	\coordinate (fc1) at (360*\s, -15*\s);
	\denseLayer{fc1}{\cd}{\s}

	\coordinate (pool2) at (310*\s, 0);
	\layer{pool2}{4*\s}{16}{\s}

	\coordinate (conv2) at (275*\s, 0);
	\layer{conv2}{8*\s}{16}{\s}

	\coordinate (pool1) at (230*\s, 0);
	\layer{pool1}{12*\s}{\cb}{\s}

	\coordinate (conv1) at (175*\s, 0);
	\layer{conv1}{24*\s}{\cb}{\s}

	\coordinate (svm) at (393*\s, 15*\s);

	\coordinate (outconverter) at (380*\s, 40*\s);

   \path[fill overzoom image=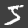] ($(input)+(-20*\s, -20*\s)$) rectangle ($(input)+(20*\s, 20*\s)$);
\path[fill overzoom image=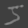, fill opacity=0.2] ($(oncell)+(-20*\s, -20*\s)$) rectangle ($(oncell)+(20*\s, 20*\s)$);
\path[fill overzoom image=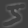] ($(offcell)+(-20*\s, -20*\s)$) rectangle ($(offcell)+(20*\s, 20*\s)$);
    \conv{$(oncell)+(6*\s,-6*\s)$}{4*\s}{\ca}{$(conv1)+(10*\s,-10*\s)$}{1*\s}{\cb}{\s}
	\conv{$(offcell)+(6*\s,-6*\s)$}{4*\s}{\ca}{$(conv1)+(10*\s,-10*\s)$}{1*\s}{\cb}{\s}
	\pool{$(conv1)+(-8*\s,16*\s)$}{2*\s}{\cb}{$(pool1)+(-4*\s,8*\s)$}{1*\s}{\cb}{\s}
	\conv{$(pool1)+(4*\s,-6*\s)$}{5*\s}{\cb}{$(conv2)+(4*\s,-6*\s)$}{1*\s}{\cc}{\s}
	\pool{$(conv2)+(-4*\s,4*\s)$}{2*\s}{\cc}{$(pool2)+(-2*\s,2*\s)$}{1*\s}{\cc}{\s}
	\dense{pool2}{\cc}{fc1}{\cd}{\s}{4}

	\draw[dashed] ($(fc1)+(\s, -\s)$) -- (svm.south -| svm.east) ;
	\draw[dashed] ($(fc1)+(-\cd*\s+\s, \cd*\s+\s)$) -- (svm.south -| svm.east) ;

	\node[draw=black, fill=white] at (svm) {SVM};

	\newcommand{\maxy}{46}
	\newcommand{\miny}{-44}

	\node (preprocess)[draw=black, dashed, fit={($(input)+(-24*\s, \maxy*\s)$)  ($(inconverter |- input)+(-3*\s,\miny*\s)$)}] {};
	\node (snn)[draw=black, dashed, fit={($(inconverter |- input)+(3*\s,\maxy*\s)$)  ($(outconverter |- input)+(-3*\s,\miny*\s)$)}] {};
	\node (classification)[draw=black, dashed, fit={($(outconverter |- input)+(3*\s, \maxy*\s)$)  ($(svm  |- input)+(9*\s,\miny*\s)$)}] {};

	\node[draw=black, fill=white] at (inconverter) {Latency coding};
	\node[draw=black, fill=white] at (outconverter) {Output conversion};

	\draw[dashed] ($(input)+(20*\s,  20*\s)$) -- ($(oncell)+(-20*\s, 20*\s)$) ;

	\draw[dashed] ($(input)+(20*\s,  20*\s)$) -- ($(oncell)+(-20*\s, 64*\s)$) ;

	\draw[dashed] ($(input)+(20*\s,  -20*\s)$) -- ($(oncell)+(-20*\s, 24*\s)$) ;

	\draw[dashed] ($(input)+(20*\s,  -20*\s)$) -- ($(oncell)+(-20*\s, -20*\s)$) ;

	\coordinate (text) at (0, -50*\s);

	\node at (input |- text) {Input};
	\node at (oncell |- text) {On/Off filters};
	\node at (conv1 |- text) {Convolution};
	\node at (pool1 |- text)  {Pooling};
	\node at (conv2 |- text)  {Convolution};
	\node at (pool2 |- text)  {Pooling};
	\node at ($(fc1 |- text)+(-10*\s, 0)$)  {Fully-connected layer};
    \node at (svm |- text)  {Classifier};

    \node[rotate=270] at ($(conv1)+(28*\s, 0)$) {$L_h(1)$};
    \node at ($(conv1)+(0, -29*\s)$) {$L_w(1)$};
    \node[rotate=-45]  at ($(conv1)+(28*\s, 28*\s)$) {$L_d(1)$};

\end{tikzpicture}}
    \caption{Network topology.}
    \label{fig:topology}
\end{figure*}

If the output layer has only one column, it directly outputs vector $y$. An SVM with a linear kernel is trained over the output training set. SVM parameters are not optimized (we set $C = 1$). Figure~\ref{fig:topology} shows the complete network topology. Besides classification rates, we investigate the sparsity of the network. The sparsity is computed over the output vectors $y$ of the test set with the following formula, used in~\cite{hoyer04a}:
\begin{equation}
\mathrm{sp}(y) = \frac{\sqrt{n_y} - \frac{\sum_i^{n_y} |y_i|}{\sqrt{\sum_i^{n_y} y_i^2}}}{\sqrt{n_y}-1}
\end{equation}
with $n_y$ the vector dimension. This measure produces values in [0, 1]. Values close to 1 mean that the vector is sparse (i.e. most of the features are close to 0).

All the results reported in this paper are averaged over 10 runs. The default parameters are reported in Table~\ref{tab:parameters}.

 \begin{table}[t]
\begin{center}
\resizebox{\columnwidth}{!}{
\begin{tabular}{|lr|lr|lr|}
    \hline
    \multicolumn{6}{|c|}{Learning}\\
    \hline
    $\lambda$ & 0.95 & $N_{\sf epoch}$ & 100 &&\\
    \hline
    \multicolumn{6}{|c|}{STDP}\\
    \hline
    $W_{\sf min}$ & 0.0 & $W_{\sf max}$ & 1.0 & $\eta_W(0)$ & 0.1\\
    $\beta$ & 1.0 & $\tau$ & 0.1 & $W(0)$ & $\sim\mathcal{U}(0, 1)$\\
    \hline
    \multicolumn{6}{|c|}{Neural Coding}\\
    \hline
    $T_{\sf start}$ & 0.0 & $T_{\sf end}$ & 1.0 &&\\
    \hline
    \multicolumn{6}{|c|}{Threshold Adaptation}\\
    \hline
	$t_{\sf target}$ & 0.7 & $\eta_{\sf Th}(0)$ & 1.0 & $\mathrm{Th}_{\sf min}$ & 1.0\\
	$V_{\sf Th}(0)$& $\sim\mathcal{N}(5, 1)$ & $V_{\sf inh}$ & 1.0 & & \\
    \hline
    \multicolumn{6}{|c|}{Pre-processing}\\
    \hline
    $\mathrm{DoG}_{\sf center}$ & 1.0 & $\mathrm{DoG}_{\sf surround}$ & 4.0 & $\mathrm{DoG}_{\sf size}$ & 7\\
    \hline
\end{tabular}
}

\end{center}
\caption{Default SNN parameters used in the experiments. $\mathcal{N}(\mu, \sigma)$ is a normal distribution centered in $\mu$ and with variance $\sigma$. $\mathcal{U}(a, b)$ is a uniform distribution in $[a, b]$.}
\label{tab:parameters}
\end{table}

\subsection{MNIST}

MNIST is a handwritten digit dataset~\cite{lecun1998gradient}. The training set contains 60,000 samples and the test set contains 10,000 samples. The network architecture is detailed in Table~\ref{tab:mnist_architecture}.

\begin{table}
\centering

\begin{tabular}{|l|c|c|c|c|}
    \hline
    Type & Filter size & Filter number & Stride & Padding\\
   \hline
    Convolution & $5 \times 5$ & 32 & 1 & 0  \\
    \hline
    Pooling & $2 \times 2$ & 32 & 2 & 0 \\
    \hline
    Convolution & $5 \times 5$ & 128 & 1 & 0  \\
    \hline
    Pooling & $2 \times 2$ & 128 & 2 & 0 \\
    \hline
    Fully-connected & $4 \times 4$ & 4096 & 1 & 0 \\
    \hline
\end{tabular}
\caption{Architecture used with the MNIST dataset.}
\label{tab:mnist_architecture}
\end{table}

\subsubsection{Threshold Target Time}

\begin{figure}[ht]
\centering

	\subfloat[$t_{\sf target} = 0.3$\label{fig:t_target_3}]{
    	\resizebox{!}{1.5cm}{
    	\begin{tabular}{cc}
    		\includegraphics{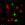} &
    		\includegraphics{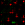} \\
   			\includegraphics{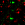} &
    		\includegraphics{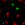} \\
    	\end{tabular}
        }
    }
	\subfloat[$t_{\sf target} = 0.5$\label{fig:t_target_5}]{
    	\resizebox{!}{1.5cm}{
        \begin{tabular}{cc}
    		\includegraphics{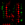} &
    		\includegraphics{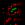} \\
   			\includegraphics{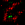} &
    		\includegraphics{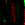} \\
    	\end{tabular}
        }
    }

    \subfloat[$t_{\sf target} = 0.7$\label{fig:t_target_7}]{
    	\resizebox{!}{1.5cm}{
        \begin{tabular}{cc}
    		\includegraphics{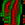} &
    		\includegraphics{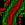} \\
   			\includegraphics{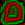} &
    		\includegraphics{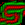} \\
    	\end{tabular}
        }
    }
    \subfloat[$t_{\sf target} = 0.9$\label{fig:t_target_8}]{
        \resizebox{!}{1.5cm}{
        \begin{tabular}{cc}
    		\includegraphics{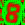} &
    		\includegraphics{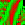} \\
   			\includegraphics{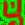} &
    		\includegraphics{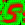} \\
    	\end{tabular}
        }
    }
    \caption{Filters learned on MNIST with multiplicative STDP w.r.t. $t_{\sf target}$.}
    \label{fig:t_target_filter}

\end{figure}

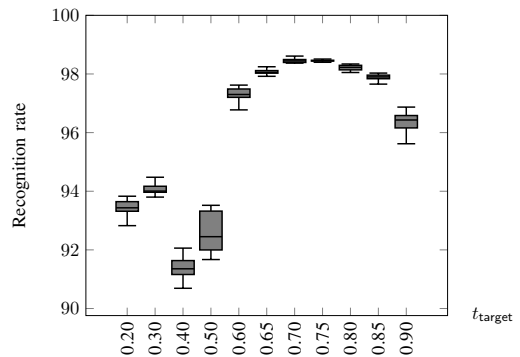
\begin{figure}
\centering
    \resizebox{0.8\columnwidth}{!}{\usepgfplotslibrary{statistics}
\pgfplotstableread{
name lw lq med uq uw 

0.20 92.82625 93.3175 93.435 93.645 93.83
0.30 93.8 93.965 94.005 94.17 94.4775
0.40 90.69 91.16 91.355 91.635 92.06
0.50 91.67 91.9975 92.45 93.3225 93.52
0.60 96.775 97.2025 97.3 97.4875 97.62
0.65 97.92 98.0275 98.055 98.115 98.24625
0.70 98.37 98.3925 98.445 98.4925 98.61
0.75 98.39875 98.44 98.45 98.4675 98.50875
0.80 98.05 98.14 98.225 98.295 98.34
0.85 97.65375 97.8375 97.9 97.96 98.03
0.90 95.62 96.16 96.43 96.5825 96.87
}\datatable

\pgfplotsset{
    boxplot prepared from table/.code={
        \def\tikz@plot@handler{\pgfplotsplothandlerboxplotprepared}%
        \pgfplotsset{
            /pgfplots/boxplot prepared from table/.cd,
            #1,
        }
    },
	boxplot/every box/.style={thick, solid, black, fill=black!50 },
	boxplot/every median/.style={thick, solid, black, fill=black!50 },
	boxplot/every whisker/.style={thick, solid, black, fill=black!50 },
    /pgfplots/boxplot prepared from table/.cd,
        table/.code={\pgfplotstablecopy{#1}\to\boxplot@datatable},
        row/.initial=0,
        make style readable from table/.style={
            #1/.code={
                \pgfplotstablegetelem{\pgfkeysvalueof{/pgfplots/boxplot prepared from table/row}}{##1}\of\boxplot@datatable
                \pgfplotsset{boxplot/#1/.expand once={\pgfplotsretval}}
            }
        },
        make style readable from table=lower whisker,
        make style readable from table=upper whisker,
        make style readable from table=lower quartile,
        make style readable from table=upper quartile,
        make style readable from table=median,
        make style readable from table=lower notch,
        make style readable from table=upper notch
}
\makeatother

\begin{tikzpicture}
\begin{axis}[
	boxplot/draw direction=y,
	ymax=100,
	xtick={1,...,11},
	xticklabels={0.20, 0.30, 0.40, 0.50, 0.60, 0.65, 0.70, 0.75, 0.80, 0.85, 0.90},
	x tick label style={rotate=90,anchor=east},
    every axis x label/.style={at={(ticklabel* cs:1.05)},anchor=west},
    xlabel={$t_{\sf target}$},
    ylabel={Recognition rate}
]
\pgfplotstablegetrowsof{\datatable}
\pgfmathtruncatemacro\TotalRows{\pgfplotsretval-1}
\pgfplotsinvokeforeach{0,...,\TotalRows}
{
  \addplot+[
  boxplot prepared from table={
    table=\datatable,
    row=#1,
    lower whisker=lw,
    upper whisker=uw,
    lower quartile=lq,
    upper quartile=uq,
    median=med
  },
  boxplot prepared,
  area legend
  ]
  coordinates {};
}
\end{axis}
\end{tikzpicture}}
    \caption{Recognition rates on MNIST according to the $t_{\sf target}$ parameter with biological STDP ($\tau = 0.1$).}
    \label{fig:t_target_result}

\end{figure}

First, we study the impact of the parameter $t_{\sf target}$. It directly impacts both the learned filters (Figure~\ref{fig:t_target_filter}) and the classification performance (Figure~\ref{fig:t_target_result}). While low values of $t_{\sf target}$ lead to very local patterns (Figure~\ref{fig:t_target_3}), larger values lead to more global patterns (Figure~\ref{fig:t_target_7}). Using late $t_{\sf target}$, and, so, training neurons to integrate a large number of spikes, helps to improve the classification rate. However, the performance decreases with very late $t_{\sf target}$: the latest spikes, which encode the lowest input values, are not useful for pattern classification. Networks with $t_{\sf target} = 0.75$ yield state-of-the-art results for SNNs trained with STDP on the MNIST dataset: 98.47\% (see Table~\ref{tab:mnist_literature} for competing approaches). The two update mechanisms described in Equation~\ref{eq:th_target} and Equation~\ref{eq:th_wta} are necessary to reach good classification rates. When Equation~\ref{eq:th_wta} is disabled in the threshold update, the homeostasis of the system is not maintained, which leads to a classification rate of $94.54\pm1.16$\% when $t_{\sf target} = 0.75$. \falez{When Equation~\ref{eq:th_target} is disabled, controlling the type of pattern to be learned becomes difficult and highly dependent on the initial values of the thresholds $V_{\sf Th}(0)$.} Using different $t_{\sf target}$ values across the layers decreases the performance (Table~\ref{tab:delta_t}). Let $\Delta_t$ be the difference between the $t_{\sf target}$ parameters of two consecutive layers. Since neurons of the previous layer are trained to fire at specific timestamps, setting an earlier $t_{\sf target}$ ($\Delta_t < 0$) on the current layer results in missing spikes from the previous neurons. Setting a later $t_{\sf target}$ ($\Delta_t > 0$) results in taking into account spikes that come too late after the $t_{\sf target}$ of the previous layer. A spike which arises too late compared to $t_{\sf target}$ means that the current pattern is not similar to those usually recognized by the input neuron. With small values of $|\Delta_t|$, the performance of the network remains stable, which shows that the threshold adaptation mechanism is noise-resistant to some extent. However, large values of $|\Delta_t|$ have a negative impact on the classification rate, especially when $\Delta_t < 0$. $\Delta_t$ is inversely proportional to the sparsity: positive values of $\Delta_t$ tend to let neurons integrate more spikes and, so, allow more neurons to fire, which decreases sparsity. For $\Delta_t = -0.20$, the classification rate and sparsity are very low because the network cannot generate any spike: the $t_{\sf target}$ of the second layer is defined to be a timestamp at which no spikes have been generated yet by the first layer. $\Delta_t=0.01$ yields the best result: 99.53\%. This small offset seems to reinforce the resistance to noise, without integrating spikes generated by unrelated patterns. These results show that finding a single value for $t_{\sf target}$ is sufficient in the exhaustive search, and the other $t_{\sf target}$ can be defined by using a very small or null $\Delta_t$. This makes it easy to set the threshold adaptation of a multilayer SNN.

\begin{table}
\centering

    \begin{tabular}{|l|c|c|}
        \hline
        $\Delta_t$ & Recognition rate & Sparsity \\
       \hline
        -0.20 & 11.35 \(\pm\) 00.00 & 0.0000 \(\pm\) 0.0000 \\
       \hline
        -0.10 & 85.56 \(\pm\) 2.28 & 0.5129 \(\pm\) 0.0230 \\
        \hline
        -0.05 & 97.68 \(\pm\) 0.14 & 0.2855 \(\pm\) 0.0067 \\
        \hline
         -0.01 & 98.36 \(\pm\) 0.05 & 0.1568 \(\pm\) 0.0068 \\
        \hline
         0.0 & 98.47 \(\pm\) 0.07 & 0.1365 \(\pm\) 0.0052 \\
         \hline
         +0.01 & 98.54 \(\pm\) 0.10  & 0.1209 \(\pm\) 0.0066 \\
         \hline
         +0.05 & 98.43 \(\pm\) 0.10 & 0.0754 \(\pm\) 0.0082 \\
         \hline
         +0.10 & 97.24 \(\pm\) 0.24 & 0.0176 \(\pm\) 0.0010 \\
         \hline
         +0.20 & 92.43 \(\pm\) 1.70 & 0.0004 \(\pm\) 0.0016 \\
        \hline
    \end{tabular}
    \caption{Results on MNIST with different $t_{\sf target}$ variations. $\Delta_t$ is the difference of $t_{\sf target}$ between consecutive layers. $t_{\sf target}$ of the first layer is fixed to $0.75$.}
    \label{tab:delta_t}
\end{table}

\subsubsection{Inhibition}

\begin{table}
\centering

    \begin{tabular}{|l|l|c|c|}
        \hline
        Inhibition policy & Layer & Recognition rate & Sparsity \\
        \hline
        \multirow{3}{*}{Winner-take-all} & Conv1 & 84.28 \(\pm\) 0.98 & 0.3389 \(\pm\) 0.0148  \\
        \cline{2-4}
        & Conv2 &  89.07 \(\pm\) 0.74 & 0.6509 \(\pm\) 0.0026 \\
        \cline{2-4}
        & FC &  61.82 \(\pm\) 1.92 &  1.0000 \(\pm\) 0.0000 \\
        \hline
        \multirow{3}{*}{Soft inhibition} & Conv1 & 85.47 \(\pm\) 0.99 & 0.2806 \(\pm\) 0.0443 \\
        \cline{2-4}
        & Conv2 & 96.14 \(\pm\) 0.68 & 0.3984 \(\pm\) 0.0171 \\
        \cline{2-4}
        & FC & 94.86 \(\pm\) 0.17 & 0.8965 \(\pm\) 0.0031 \\
        \hline
        \multirow{3}{*}{No inhibition} & Conv1 & 84.71 \(\pm\) 1.04  &  0.1538 \(\pm\) 0.0069 \\
        \cline{2-4}
        & Conv2 & 96.15 \(\pm\) 0.17 & 0.1621 \(\pm\) 0.0056 \\
        \cline{2-4}
        & FC & 98.47 \(\pm\) 0.07 & 0.1365 \(\pm\) 0.0052 \\
        \hline
    \end{tabular}
    \caption{Recognition rates on MNIST with the different inhibition policies for $t_{\sf target} = 0.75$ and biological STDP ($\tau = 0.1$).}
    \label{tab:inhibition}

\end{table}

We run experiments to show the impact of the inhibition strategy on recognition rates. We compare the three inhibition policies detailed in Section~\ref{ssec:competition}. Table~\ref{tab:inhibition} shows that increasing the hardness of inhibition during inference tends to decrease the recognition rate. This can be related to the sparsity level. The effect of inhibition, which is minimal in the first layer, is accentuated after each layer. This effect strongly impacts both the sparsity and the recognition rate in the fully connected layer. This effect is visible with soft inhibition, but is maximal with the WTA policy: the sparsity of the fully-connected layer is 1, while the recognition rate is only 63.43\%. Maintaining higher levels of activity helps to learn better representations.

\subsubsection{STDP Rule}

\begin{figure}[ht]
\centering
	\subfloat[Additive STDP\label{fig:conv1_stdp_lin}]{
    	\resizebox{0.3\columnwidth}{!}{
    	\begin{tabular}{cc}
    		\includegraphics{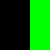} &
    		\includegraphics{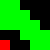} \\
   			\includegraphics{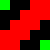} &
    		\includegraphics{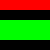} \\
    	\end{tabular}
        }
    }
	\subfloat[Multiplicative STDP\label{fig:conv1_stdp_mul}]{
    	\resizebox{0.3\columnwidth}{!}{
    	\begin{tabular}{cc}
    		\includegraphics{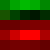} &
    		\includegraphics{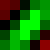} \\
   			\includegraphics{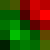} &
    		\includegraphics{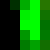} \\
    	\end{tabular}
        }
    }
	\subfloat[Biological STDP\label{fig:conv1_stdp_bio}]{
    	\resizebox{0.3\columnwidth}{!}{
        \begin{tabular}{cc}
    		\includegraphics{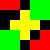} &
    		\includegraphics{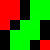} \\
   			\includegraphics{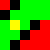} &
    		\includegraphics{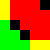} \\
    	\end{tabular}
        }
    }
    \caption{Filters in the first convolution w.r.t. the STDP rule.}
    \label{fig:conv1_stdp}

\end{figure}

We study the effects of the STDP rules on the network classification rates and sparsity. We test the three STDP rules described in Section~\ref{ssec:synapses}: additive STDP, multiplicative STDP, and biological STDP (Table~\ref{tab:stdp}). Additive STDP yields a baseline performance of 96.10\% and a relatively high level of sparsity (0.8057). Figure~\ref{fig:conv1_stdp_lin} shows that this rule leads to binary weights (0 or 1) due to a saturation effect. Multiplicative STDP reduces this effect using the $\beta$ parameter: large values of $\beta$ reduce drastically the number of weights close to 0 or 1 (Figure~\ref{fig:conv1_stdp_mul}). Table~\ref{tab:stdp} shows that increasing $\beta$ decreases the sparsity. $\beta = 3.0$ provides a classification rate of 98.22\% and a sparsity of 0.3215. Finally, the best performance (98.47\%) is reached with biological STDP with $\tau = 0.1$. Decreasing this parameter also reduces the sparsity. Figure~\ref{fig:conv1_stdp_bio} shows that filters learned by biological STDP look different from the ones learned by other STDP rules. Indeed, additive and multiplicative STDP rules never learn patterns in which the on and off channels overlap (i.e. red and green pixels are always separated in the filters), because our input coding does not allow \marius{generating} a spike from both channels at the same position. In contrast, biological STDP leads to filters with reinforced connections on the two channels (yellow pixels), which means that biological STDP is able to combine multiple patterns. Whatever the STDP rule, multiplicative or biological STDP, networks with the lowest levels of sparsity never yield the best classification performances.

\begin{table}
\centering

    \begin{tabular}{|l|c|c|}
        \hline
        STDP rule & Recognition rate & Sparsity \\
       \hline
        Additive STDP & 96.10 \(\pm\) 0.33 & 0.8057 \(\pm\) 0.0127 \\
        \hline
        Multiplicative STDP ($\beta = 2.0$)& 97.99 \(\pm\) 0.10 & 0.6298 \(\pm\) 0.0052 \\
        \hline
        Multiplicative STDP ($\beta = 3.0$)& 98.22 \(\pm\) 0.06 & 0.3215 \(\pm\) 0.0154 \\
        \hline
        Multiplicative STDP ($\beta = 4.0$)& 97.67 \(\pm\) 0.11 & 0.1203 \(\pm\) 0.0044 \\
        \hline
        Biological  STDP ($\tau = 0.05$)& 98.04 \(\pm\) 0.14 & 0.0622 \(\pm\) 0.0072 \\
        \hline
        Biological  STDP ($\tau = 0.1$)& 98.47 \(\pm\) 0.07 & 0.1335 \(\pm\) 0.0066\\
        \hline
        Biological  STDP ($\tau = 0.5$)& 98.16 \(\pm\) 0.13 & 0.2220 \(\pm\) 0.0096 \\
        \hline
    \end{tabular}
    \caption{Recognition rates on MNIST w.r.t. STDP rules ($t_{\sf target} = 0.75$).}
    \label{tab:stdp}

\end{table}

\begin{figure}[ht]
\centering

	\subfloat[Additive STDP\label{fig:conv3_stdp_mul}]{
    	\resizebox{!}{1.5cm}{
    	\begin{tabular}{cc}
    		\includegraphics{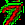} &
    		\includegraphics{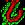} \\
   			\includegraphics{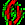} &
    		\includegraphics{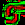} \\
    	\end{tabular}
        }
    }
	\subfloat[Biological STDP\label{fig:conv3_stdp_bio}]{
    	\resizebox{!}{1.5cm}{
        \begin{tabular}{cc}
    		\includegraphics{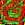} &
    		\includegraphics{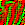} \\
   			\includegraphics{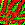} &
    		\includegraphics{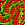} \\
    	\end{tabular}
        }
    }
    \caption{Filter reconstructions of units of fully connected layers learned with different STDP rules.}
    \label{fig:conv3_stdp}

\end{figure}

The shapes of the filters in the fully-connected layer also differ from one STDP rule to another. While additive and multiplicative STDPs lead to easily identifiable digits (Figure~\ref{fig:conv3_stdp_mul}), biological STDP provides less obvious filters (Figure~\ref{fig:conv3_stdp_bio}). The non-linearity brought by biological STDP seems to allow learning more complex features, improving performances.

\subsubsection{Multiple Target Timestamp Networks}

Finally, we use several groups of neurons with different $t_{\sf target}$. Representations learned with different target timestamps can contain more diverse patterns. We train independent networks where all layers are set with a given $t_{\sf target}$ value. Then, we merge the features at the output of each network by concatenating them and feed the resulting feature vector to the classifier. To make a fair comparison, we compare configurations that result in feature vectors of the same dimension (4096).

\begin{table}
\centering
    \resizebox{\columnwidth}{!}{
    \begin{tabular}{|l|l|c|}
        \hline
        $N$  & $t_{\sf target}$ & Rec. rate \\
        \hline
        4096 & 0.750 & 98.47 \(\pm\) 0.07\\
        \hline
        2048 &  0.300, 0.750  & 98.51 \(\pm\) 0.06 \\
        \hline
        2048 & 0.650, 0.750, & 98.53 \(\pm\) 0.06 \\
        \hline
        1024 & 0.300, 0.500, 0.700, 0.800 &  98.59 \(\pm\) 0.06 \\
        \hline
        1024 & 0.650, 0.700, 0.750, 0.800 & \textbf{98.60 \(\pm\) 0.08}\\
        \hline
        512 & 0.200, 0.300, 0.400, 0.500, 0.600, 0.700, 0.800, 0.900 & 98.48 \(\pm\) 0.05 \\
        \hline
        512 & 0.675, 0.700, 0.725, 0.750, 0.775, 0.800, 0.825, 0.850 & 98.57 \(\pm\) 0.08\\
        \hline
    \end{tabular}
    }
    \caption{Recognition rates of multi-$t_{\sf target}$ SNNs on MNIST. Each configuration has a total of 4096 output neurons.}
    \label{tab:multi_target}

\end{table}

Table~\ref{tab:multi_target} shows that using multiple targets improves the classification performance. The network reaches a recognition rate of 98.60\%, which is better than existing comparable methods (Table~\ref{tab:mnist_literature}). One explanation can be that the combination of different $t_{\sf target}$ allows detecting more varied patterns. Only methods that convert ANNs to SNNs after training~\cite{diehl2015fast} outperform our model.

\begin{table}
\centering

    \begin{tabular}{|l|l|c|}
        \hline
        Model & Description & Recognition rate \\
        \hline
        Querlioz et al. 2011\cite{querlioz2011simulation} & Single layer SNN & 93.50\\
        \hline
        Dielh et al. 2015\cite{diehl2015unsupervised} & Single layer SNN & 95.00\\
        \hline
        Tavanaei et al. 2016\cite{tavanaei2016bio} & CSNN+SVM & 98.36\\
        \hline
        Kheradpisheh et al. 2018\cite{kheradpisheh2018stdp} & CSNN+SVM & 98.40\\
        \hline
        Dielh et al. 2015\cite{diehl2015fast} & Converted CSNN & 99.10\\
        \hline
        \textbf{This work} & CSNN+SVM & \textbf{98.60}\\
        \hline
    \end{tabular}
    \caption{Comparison with different spiking models with STDP from the literature (MNIST).}
    \label{tab:mnist_literature}

\end{table}

\subsection{Faces/Motorbikes}

Finally, we test our model on the Faces/Motorbikes dataset used in~\cite{kheradpisheh2018stdp} in order to ensure that the model performs well with more realistic images. The dataset contains two classes extracted from the Caltech-101 dataset: faces and motorbikes. Similarly to~\cite{kheradpisheh2018stdp}, images are resized to $250\times 160$ pixels, then converted into the grayscale format. The training set has 474 samples and the test set has 759 samples. Since our training protocol differs from~\cite{kheradpisheh2018stdp} (Section~\ref{ssec:training}), it is necessary to increase the number of filters in the convolution layer and to use larger values for $\mathrm{Th}_{\sf min}$ (in the following experiment, we choose 8) to focus on patterns resulting from enough spikes. We use additive STDP in all the convolution layers. The detailed architecture is provided in Table~\ref{tab:face_motor_architecture}.

\begin{table}
\centering

\begin{tabular}{|l|c|c|c|c|}
    \hline
    Type & Filter size & Filter number & Stride & Padding\\
   \hline
    Convolution & $5 \times 5$ & 32 & 1 & 2  \\
    \hline
    Pooling & $7 \times 7$ & 32 & 6 & 3 \\
    \hline
    Convolution & $17 \times 17$ & 64 & 1 & 8  \\
    \hline
    Pooling & $5 \times 5$ & 64 & 5 & 2 \\
    \hline
    Convolution & $5 \times 5$ & 128 & 1 & 2 \\
    \hline
\end{tabular}
\caption{Architecture used on Faces/Motorbikes.}
\label{tab:face_motor_architecture}
\end{table}

\begin{figure}
  \begin{center}
  	\resizebox{0.8\columnwidth}{!}{\usepgfplotslibrary{statistics}
\pgfplotstableread{
name lw lq med uq uw 

0.2 96.8379 97.924925 98.41895 98.847175 99.6047
0.3 97.6285 98.320175 98.946 99.17655 99.6047
0.4 97.7602 98.32015 98.419 98.781275 98.946
0.5 97.365 97.6285 97.8261 98.1555 98.946
0.6 98.0237 98.386025 98.74835 98.8142 98.946
0.7 97.6285 98.089575 98.48485 98.64955 99.0777
0.8 99.0777 99.275375 99.473 99.6047 99.8682
0.9 98.1555 98.1884625 98.2872 98.748325 99.0777

}\datatable

\pgfplotsset{
    boxplot prepared from table/.code={
        \def\tikz@plot@handler{\pgfplotsplothandlerboxplotprepared}%
        \pgfplotsset{
            /pgfplots/boxplot prepared from table/.cd,
            #1,
        }
    },
	boxplot/every box/.style={thick, solid, black, fill=black!50 },
	boxplot/every median/.style={thick, solid, black, fill=black!50 },
	boxplot/every whisker/.style={thick, solid, black, fill=black!50 },
    /pgfplots/boxplot prepared from table/.cd,
        table/.code={\pgfplotstablecopy{#1}\to\boxplot@datatable},
        row/.initial=0,
        make style readable from table/.style={
            #1/.code={
                \pgfplotstablegetelem{\pgfkeysvalueof{/pgfplots/boxplot prepared from table/row}}{##1}\of\boxplot@datatable
                \pgfplotsset{boxplot/#1/.expand once={\pgfplotsretval}}
            }
        },
        make style readable from table=lower whisker,
        make style readable from table=upper whisker,
        make style readable from table=lower quartile,
        make style readable from table=upper quartile,
        make style readable from table=median,
        make style readable from table=lower notch,
        make style readable from table=upper notch
}
\makeatother

\begin{tikzpicture}
\begin{axis}[
	boxplot/draw direction=y,
    xmin=0,
    xmax=9,
	ymax=100,
    ymin=96.5,
	xtick={1,...,8},
	xticklabels={0.2, 0.3, 0.4, 0.5, 0.6, 0.7, 0.8, 0.9},
	x tick label style={rotate=90,anchor=east},
    every axis x label/.style={at={(ticklabel* cs:1.05)},anchor=west},
    enlargelimits=false,
    legend entries={Baseline},
    legend pos = south east,
    xlabel={$t_{\sf target}$},
    ylabel={Recognition rate}
]
\addlegendimage{line legend, red, dashed}
\pgfplotstablegetrowsof{\datatable}
\pgfmathtruncatemacro\TotalRows{\pgfplotsretval-1}
\pgfplotsinvokeforeach{0,...,\TotalRows}
{
  \addplot+[
  boxplot prepared from table={
    table=\datatable,
    row=#1,
    lower whisker=lw,
    upper whisker=uw,
    lower quartile=lq,
    upper quartile=uq,
    median=med
  },
  boxplot prepared,
  area legend
  ]
  coordinates {};
}
\addplot[domain=0:9,red, dashed] (x, 99.1);

\end{axis}
\end{tikzpicture}}
  \end{center}
  \caption{Recognition rates on Faces/Motorbikes according to the $t_{\sf target}$ used. The baseline is the best result reported in~\cite{kheradpisheh2018stdp}.}
  \label{fig:face_motor_result}
\end{figure}
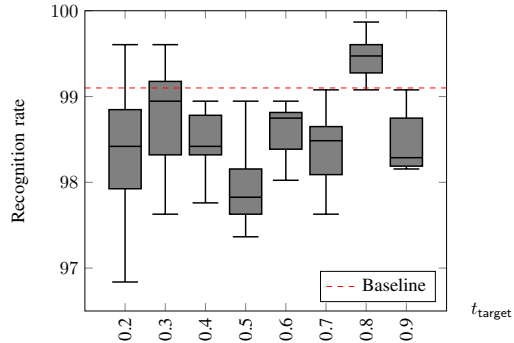

\begin{figure}[ht]
	\begin{center}
	\subfloat[Convolution 1]{
    	\resizebox{.45\columnwidth}{!}{
    	\begin{tabular}{cc}
    		\includegraphics[width=1cm]{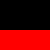} &
    		\includegraphics[width=1cm]{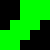} \\
   			\includegraphics[width=1cm]{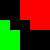} &
    		\includegraphics[width=1cm]{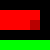} \\
    	\end{tabular}
        }
    }
    \subfloat[Convolution 2]{
    	\resizebox{.45\columnwidth}{!}{
    	\begin{tabular}{cc}
    		\includegraphics[width=1cm]{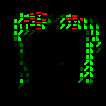} &
    		\includegraphics[width=1cm]{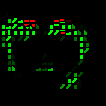} \\
   			\includegraphics[width=1cm]{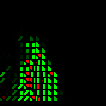} &
    		\includegraphics[width=1cm]{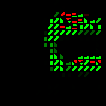} \\
    	\end{tabular}
        }
    }

    \subfloat[Convolution 3]{
    	\resizebox{0.95\columnwidth}{!}{
    	\begin{tabular}{cccc}
    		\includegraphics[width=1cm]{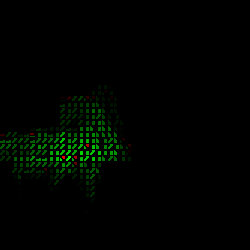} &
    		\includegraphics[width=1cm]{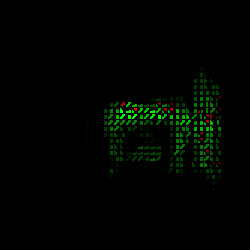} &
   			\includegraphics[width=1cm]{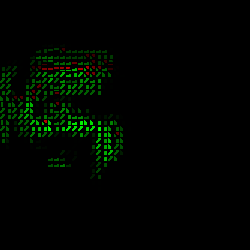} &
    		\includegraphics[width=1cm]{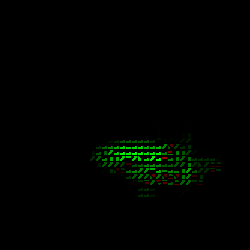} \\
    	\end{tabular}
        }
    }
	\end{center}
    \caption{Reconstruction of the receptive fields of filters learned on Faces/Motorbikes in the different layers.}
    \label{fig:face_motor_filter}
\end{figure}

Our model gives results similar to those reported in~\cite{kheradpisheh2018stdp} (Figure~\ref{fig:face_motor_result}), where the best reported result is 99.1\%. When using  $t_{\sf target} = 0.8$, our model performs better with an average of 99.46\%. The learned filters are similar to~\cite{kheradpisheh2018stdp} (Figure~\ref{fig:face_motor_filter}).

\section{Discussion}

Our model is almost fully local and is unsupervised from the input to the classifier. However, convolution layers remain an issue for hardware implementations. We succeeded in learning one convolution column independently from the others, but the weight and threshold values still have to be copied onto the other columns  after training. This is needed to reconstruct the geometry of the feature maps, to apply pooling for instance. Moreover, we used a linear SVM for classification. However, to have a fully hardware-implementable SNN, using bio-inspired classifiers is required. Recent work succeeds in using supervised STDP as a classifier in multi-layered SNN~\cite{mozafari2018first}. We aim at investigating the performance of our model with such learning rules, while respecting the constraint of local computations. Finally, results showed that $t_{\sf target}$  has a strong impact on the classification performance of the network. This parameter could be made auto-adaptable, so that neurons can find by themselves the best timing for firing. Such mechanisms would have the advantage of setting an optimal $t_{\sf target}$ value for each feature independently.

\section{Conclusion}

Previous multi-layered SNN models require a particular attention in setting neuron thresholds. An exhaustive search is needed to optimize them. Moreover, the optimal values vary from one layer to another~\cite{kheradpisheh2018stdp}. We introduced a threshold adaptation mechanism, which relies on a single parameter for all the layers and learns more varied patterns. Experiments showed that our model leads to state-of-the-art results with unsupervised SNNs on MNIST (98.60\%) and on Faces/Motorbikes (99.46\%).  We also showed that removing the inhibition during the inference step helps to reduce the sparsity of the model, which leads to an improvement of the performance. We also investigated the STDP rules and showed that biological STDP helps to improve the network performance by introducing non-linearity.

\bibliographystyle{plain}
\balance
\bibliography{main}

\end{document}